\documentclass[preprint]{acm_proc_article-sp}

\newcommand{\UPLB}{U.P. Los Ba\~{n}os}

\let\footnotesize\small

\begin{document}

\title{A Neural Prototype for a Virtual Chemical Spectrophotometer}
\numberofauthors{3}
\author{
\alignauthor Jaderick P. Pabico\\
   \affaddr{Institute of Computer Science}\\
   \affaddr{\UPLB}\\
   \email{jppabico@up.edu.ph}
\alignauthor Jose Rene L. Micor\\
   \affaddr{Institute of Chemistry}\\
   \affaddr{\UPLB}\\
   \email{jrlmicor@up.edu.ph}
\alignauthor Elmer Rico E. Mojica\\
   \affaddr{Department of Chemistry}\\
   \affaddr{State University of New York}\\
   \email{elmericomojica@gmail.com}
}
\date{}
\maketitle

\begin{abstract}
A virtual chemical spectrophotometer for the simultaneous analysis of nickel (Ni) and cobalt (Co) was developed based on an artificial neural network (ANN). The developed ANN correlates the respective concentrations of Co and Ni given the absorbance profile of a Co-Ni mixture based on the Beer’s Law. The virtual chemical spectrometer was trained using a 3-layer jump connection neural network model (NNM) with 126 input nodes corresponding to the 126 absorbance readings from 350 nm to 600 nm, 70 nodes in the hidden layer using a logistic activation function, and 2 nodes in the output layer with a logistic function. Test result shows that the NNM has correlation coefficients of 0.9953 and 0.9922 when predicting [Co] and [Ni], respectively. We observed, however, that the NNM has a duality property and that there exists a real-world practical application in solving the dual problem: Predict the Co-Ni mixture's absorbance profile given [Co] and [Ni]. It turns out that the dual problem is much harder to solve because the intended output has a much bigger cardinality than that of the input. Thus, we trained the dual ANN, a 3-layer jump connection nets with 2 input nodes corresponding to [Co] and [Ni], 70-logistic-activated nodes in the hidden layer, and 126 output nodes corresponding to the 126 absorbance readings from 250 nm to 600 nm. Test result shows that the dual NNM has correlation coefficients that range from 0.9050 through 0.9980 at 356 nm through 578 nm with the maximum coefficient observed at 480 nm. This means that the dual ANN can be used to predict the absorbance profile given the respective Co-Ni concentrations which can be of importance in creating academic models for a virtual chemical spectrophotometer.
\end{abstract}


\section{Introduction}\label{intro}
An artificial neural network (ANN) is a system loosely based or modeled on the human brain. It can goes by many names, such as natural intelligent system, connectionism, neuron computing, parallel distributed processing, machine learning algorithms and artificial neural networks~\cite{ref1}. The ANN is able to obtain or attain information and offer or present models even when the information and data are complex, noise contaminated, nonlinear and incomplete~\cite{ref2,ref3,ref4,ref5}. 

Artificial neural networks (ANNs) are powerful tools, exceptionally suited for a diverse tasks in information processing, such as recognizing patterns, generalizing, analyzing non-linear multivariate data and other things~\cite{ref4,ref5}. Currently, the most commonly used ANN type is a multi-layer feed forward network, which is trained by the back propagation (BP) algorithm. The application of ANNs to data is claimed to constitute so-called ``soft models,'' since the models have an ability to learn and extract $X$--$Y$ relationships from the presentation of a set of training samples. Their flexibility has been a decisive asset compared with parametric techniques that require the assumption of a specific hard model form. In addition, ANNs avoid the time-consuming and possibly expensive task of hard model identification~\cite{ref4}. There are many applications of ANNs particularly in fields like medicine~\cite{ref6}, engineering, chemistry, physics, agriculture, music, economy and management, archeology, as well as industry~\cite{ref7,ref8,ref9,ref10,ref11,ref12,ref13,ref14}. In chemistry, ANNs have the ability to tackle the problem of complex relationships among variables that cannot be accomplished by more traditional methods. The ANN modeling method has found extensive application in the field of simultaneous determination of several species in a given sample. This method makes it possible to eliminate or reduce the effect of the analyte-analyte interaction, the multi-step process and any other unknown non-linearity in systems~\cite{ref15}. Another advantage of ANN is its anti-jamming, anti-noise and robust nonlinear transfer ability which in a proper model would results in lower calibration errors and prediction errors~\cite{ref16}.

Nickel (II) and cobalt (II) are metals that appear together in many real samples, both natural and artificial.  Several techniques such as atomic absorption, atomic fluorescence, X-ray fluorescence, voltammetric and spectrophotometric methods have been used for the determination of these ions in different samples. Among the most widely used analytical methods are those based on the UV\_visible spectrophotometric techniques due to the resulting experimental rapidly, simplicity and the wide application~\cite{ref17}.

Simultaneous determination of trace amounts of metals in environmental samples is still a challenging analytical problem because of the sensitivity and specificity required in environmental monitoring and regulations. Recently, spectrophotometric methods based on ANNs have found increasing applications for multicomponent determination. This method is effective because they can improve the performance and application of the analytical method with the use of simultaneous analysis of several spectra. There are several studies that reported on the simultaneous analysis of metals using ANN. An example of this is the study on the simultaneous spectrophotometric determination of Co(II) and Ni(II) based on formation of their complexes with EDTA complexes with pyrolidine and carbon disulfide~\cite{ref17}. It uses ANN to analyze the mixture spectra of the complex solutions formed. In this paper, mixture of untreated Ni(II) and Co (II) were prepared and the concentration of each component was determined based on the absorbance profile of the standard solutions.

\section{Methodology}
\subsection{Experimental}

All chemicals were of analytical reagent grade and deionized water was used throughout the experiment. Stock standard solution (0.50 M) of Ni(II) and Co(II) were prepared by dissolving appropriate amounts of nickel nitrate (Sigma) and cobalt nitrate (Sigma), respectively in 25 mL volumetric flasks and diluted to the mark with deionized water.

\subsection{Procedure}

Sample solutions ranging from 0.02 to 0.10 M of Ni(II) and Co(II) were prepared in 5 mL volumetric flask by taking a required volume of the stock solution and then diluted to the mark with deionized water.  A mixture was also prepared ranging with either one of the component containing a concentration that ranged from 0.025 to 0.10 M.

\subsection{Instrumentation}

Quartz cuvettes (1 cm$^-2$) were used for all spectroscopic experiments and all measurements were performed at room temperature (25 $\pm$ 1C). All absorbance measurements were carried out on a Hewlett Packard 8452A diode array spectrophotometer with a 1 nm spectral bandpass. For each concentration of each metal and the mixture, the spectrum was scanned in the wavelength of 350--600 nm. 

\subsection{Training Data and Randomization}

The data collected from the spectrophotometer totaled 6,000 records. To facilitate input into the ANN, the absorbance data were normalized to values between~0 and~1. A randomization function was created that fed randomized data into the ANN during the training and testing phases.

\subsection{ANN Optimization by Artificial Chemistry}

Artificial Chemistry (AChem) is a computational paradigm for search, optimization, and machine learning. In this paradigm, the artificial molecules represent machine or data and the interactions among these molecules are driven by an algorithm. The duality of the molecule to represent either a machine (or operator) or data (or operand) enables a molecule to process other molecules or be processed. This dualism property of molecules enables one to implicitly define a constructive computational procedure using the dynamics of chemical reaction as a metaphor to solve complex optimization problems~\cite{pabico03}. The capabilities of AChem to simultaneously find solutions to different problems makes it a potent solution to finding the best ANN structure for this problem.

An AChem system was setup to simultaneously find the optimal ANN structure. The AChem system ran for 10,000 simulation cycles when 80\% of the molecules already encode the same ANN structure. Each AChem molecule encodes the number of hidden layers, the number of neurons on each hidden layer, a binary digit that flags whether the structure will use a jump-connection or not, the learning rate, and the momentum value. The number of neurons in the input and output layers are fixed at 2 and 126, respectively. The 2 input neurons correspond to the [Co] and [Ni] while the 126 output neurons correspond to the absorbance readings from 250 nm to 600 nm. Specialized reaction rules were devised such that the collision of the two molecules, as well as the collision of a molecule with the artificial reaction tank, will create more molecules. A reactor algorithm was also devised to filter out molecules that encode better solutions~\cite{pabico07}.

Each encoded ANN structure on AChem's molecule was evaluated by running a feed-forward, back-propagation learning algorithm over the training data using the accompanying encoded learning rate and momentum. The ANN used 4,200 records as the training set. A test set, disjoint from the training set, was extracted containing 600 records. The ANN training was stopped when the network error over the test set has not improved after 100 epochs. The trained ANN was when the network error over the test set was at the minimum. A validation set, disjoint from both the training and test sets, was extracted containing 1,200 records. The trained ANN was run over the validation set and its classification rate recorded. The recorded classification rate becomes the molecule's evaluation.

\section{Results and Discussion}

\begin{figure*}[p]
\centering
\epsfig{file=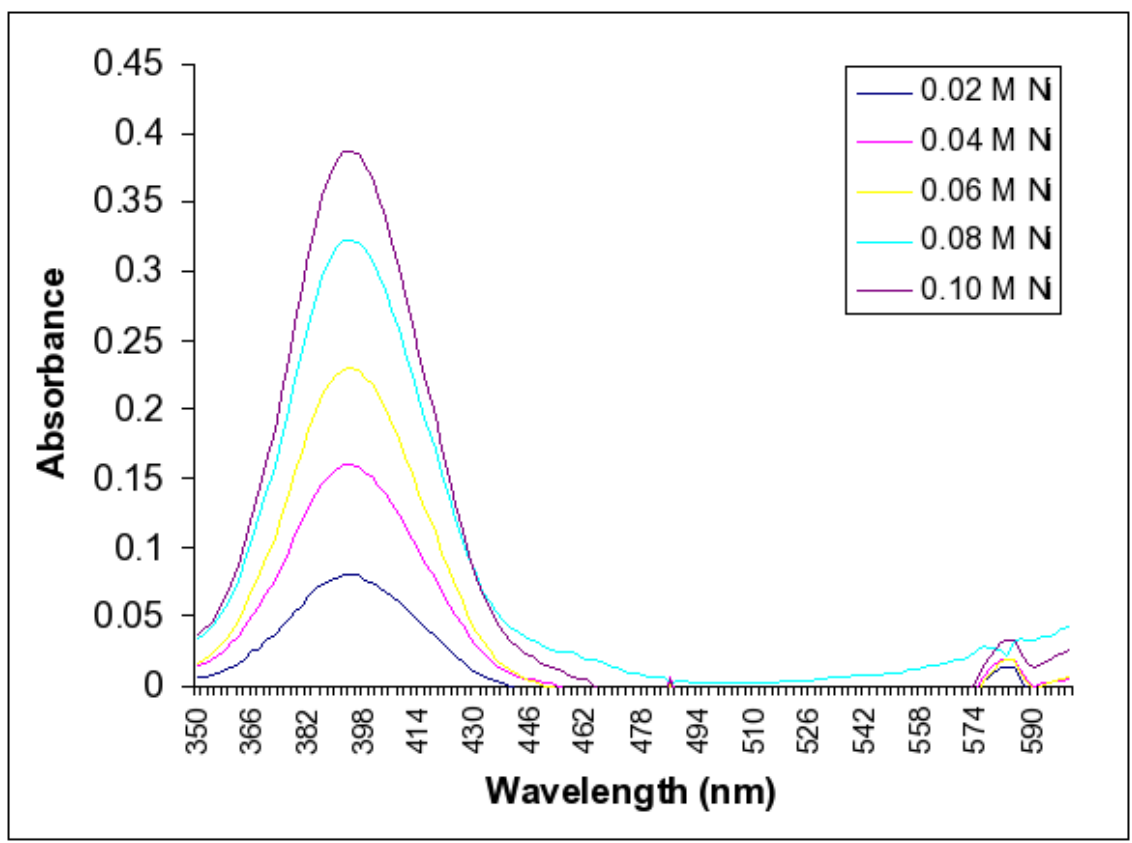, width=6in,height=4.46in}
\caption{Absorbance profile of nickel solution at increasing concentration.}
\label{fig1}
\end{figure*}

\begin{figure*}[p]
\centering
\epsfig{file=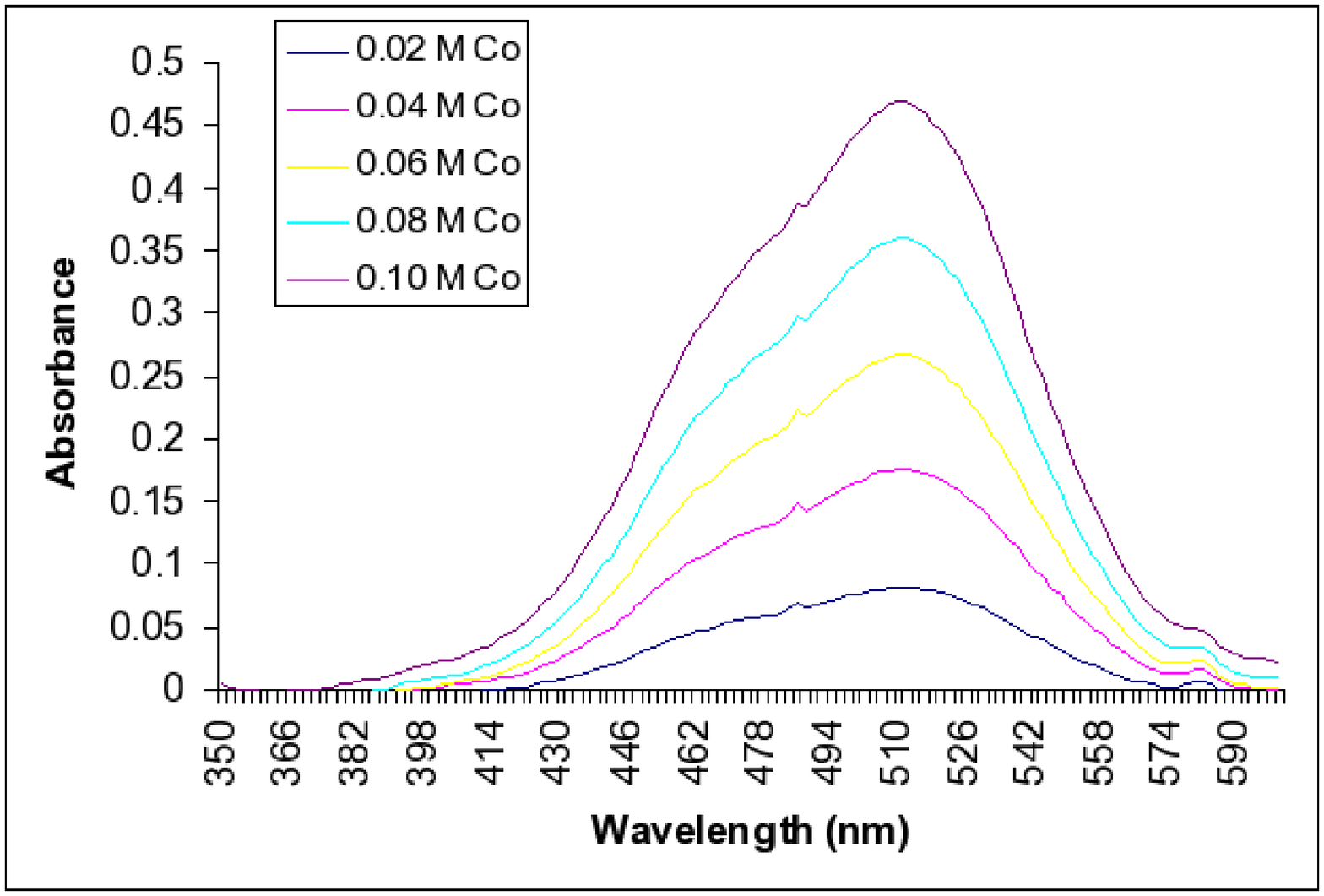, width=6in,height=4.08in}
\caption{Absorbance profile of cobalt solution at increasing concentration.}
\label{fig2}
\end{figure*}

\begin{figure*}[p]
\centering
\epsfig{file=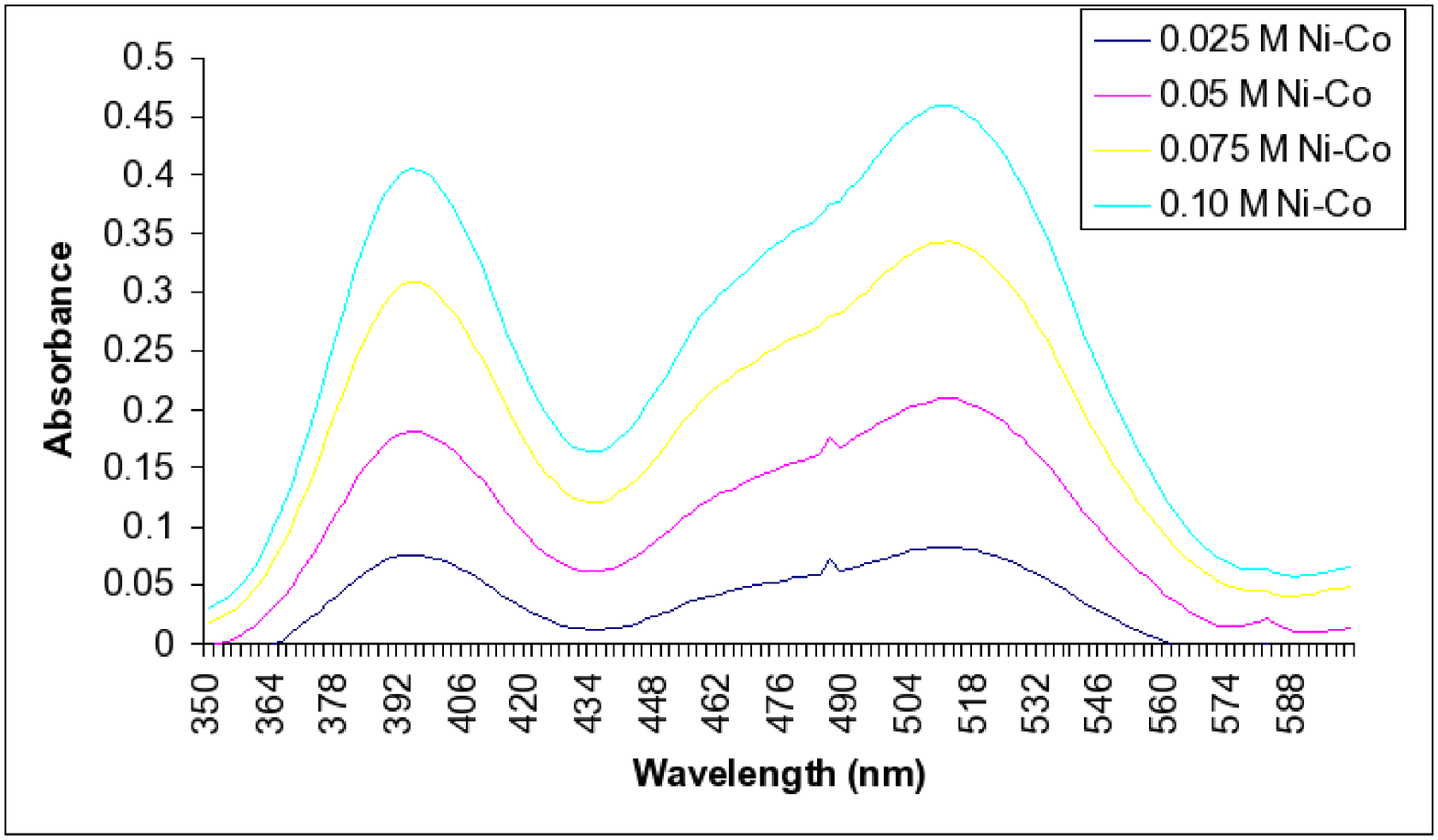, width=6in,height=3.52in}
\caption{Absorbance profile of a mixed nickel and cobalt solution at increasing concentration.}
\label{fig3}
\end{figure*}

The absorbance profile of the solutions containing nickel, cobalt and both nickel and cobalt were obtained. Figures~\ref{fig1}-\ref{fig3} showed the increase of absorbance at increasing concentration of the given solution. The increase in absorbance is directly proportional to the increase in concentration. Ni solution has an absorbance maximum at around 394 nm while the cobalt solution had an absorbance maximum at around 510 nm. These maxima was also observed in the absorbance profile of the solution containing both metals. Although overlapping of the spectrum was observed, the absorbance maxima is still proportional to the concentration of the metal. With this result, a direct relationship correlating the concentration of metal in the solution with the absorbance was obtained and used to develop an artificial neural network which can predict the concentration of the components in a given solution.

The AChem optimization routine found the optimal ANN structure as a 3-layer jump connection nets with 2 input nodes corresponding to [Co] and [Ni], 70-logistic-activated nodes in the hidden layer, and 126 output nodes corresponding to the 126 absorbance readings from 250 nm to 600 nm. Test result shows that the dual NNM has correlation coefficients that range from 0.9050 through 0.9980 at 356 nm through 578 nm with the maximum coefficient observed at 480 nm. This means that the dual ANN can be used to predict the absorbance profile given the respective Co-Ni concentrations which can be of importance in creating academic models for a virtual chemical spectrophotometer.

\bibliography{paper}
\bibliographystyle{abbrv}
\balancecolumns
\end{document}